\documentstyle[colap]{article}

\newcommand{\bdes}{\begin{description}}
\newcommand{\edes}{\end{description}}
\newcommand{\bite}{\begin{itemize}}
\newcommand{\eite}{\end{itemize}}
\newcommand{\benum}{\begin{enumerate}}
\newcommand{\eenum}{\end{enumerate}}
\newcommand{\bverb}{\begin{verbatim}}
\newcommand{\everb}{\end{verbatim}}
\newcommand{\m}[1]{{\em #1\/}}

\newcounter{thesubeg}
\newcounter{thesubegfoo}

\newenvironment{eg}[1]{\refstepcounter{equation}\label{#1}\samepage\begin{list}
{(\arabic{equation})}{\usecounter{thesubeg}}}{\end{list}}

\newcommand{\refeg}[1]{(\ref{#1})}

\title{\submitted{Appears in {\em Proceedings of the 16th
International Conference on Computational Linguistics (COLING-96)},
Copenhagen, 5--9 August 1996}
A Corpus Study of Negative Imperatives in \\ Natural Language
Instructions\thanks{\hspace*{1mm} This work is partially supported by the
Engineering and Physical Sciences Research Council ({\sc epsrc}) Grant
\mbox{J19221} and by the Commission
of the European Union Grant \mbox{{\sc lre}-62009}.}}
\author{Keith Vander Linden\thanks{\hspace*{1mm} After September 1, Dr. Vander Linden's
address will be Dept. of Mathematics and Computer Science, Calvin
College, Grand Rapids, MI 49546, USA.}\\
        Information Technology Research Institute\\
	University of Brighton\\
	Brighton BN2 4AT, UK\\
	knvl@itri.brighton.ac.uk\\
	\And
	Barbara Di Eugenio\\
	Computational Linguistics\\
	Carnegie Mellon University\\
	Pittsburgh, PA, 15213  USA\\
	dieugeni@andrew.cmu.edu\\
	}

\begin{document}

\makeatletter
 \def\submitted#1{\setbox\@tempboxa\vbox{\normalsize \tt \raggedright
    #1 \\ \hbox{}}
    \vspace{-1.5 cm} \usebox\@tempboxa \\
    \vspace{-\ht\@tempboxa} \vspace{1.5 cm}}
\makeatother

\maketitle

\begin{abstract}
In this paper, we define the notion of a preventative
expression and discuss a corpus study of such expressions in
instructional text.  We discuss our coding schema, which takes into
account both form and function features, and present measures of
inter-coder reliability for those features. We then discuss the
correlations that exist between the function and the form features.  
\end{abstract}

\section{Introduction}

While interpreting instructions, an agent is continually faced with a
number of possible actions to execute, the majority of which are not
appropriate for the situation at hand.  An instructor is therefore
required not only to prescribe the appropriate actions to the reader,
but also to prevent the reader from executing the inappropriate and
potentially dangerous alternatives.  The first task, which is commonly
achieved by giving simple imperative commands and statements of
purpose, has received considerable attention in both the
interpretation (e.g., \cite{thesis}) and the generation communities
(e.g., \cite{vanlin95}).  The second, achieved through the use of
\m{preventative expressions}, has received considerably less
attention.  Such expressions can indicate actions that the agent
should not perform, or manners of execution that the agent should not
adopt.  An agent may be told, for example, ``Do not enter'' or ``Take
care not to push too hard''.

Both of the examples just given involve negation (``do \m{not}'' and
``take care \m{not}'').  Although this is not strictly necessary for
preventative expressions (e.g., one might say ``stay out'' rather than
``do not enter''), we will focus on the use of negative forms in this
paper.  We will use the following categorisation of explicit
preventative expressions:

\bite
\item  negative imperatives proper (termed \m{DONT} imperatives). These are
characterised by the negative auxiliary {\em do not} or {\em don't}.
\begin{eg}{vinyl-asb}
\item \m{Your sheet vinyl floor may be vinyl asbestos, which is  no longer
on the market. {\bf Don't sand it or tear it up}
because this will put 
dangerous asbestos fibers into the air.}
\end{eg}
\item other preventative imperatives (termed \m{neg-TC}
imperatives). These include {\em take care} and {\em be careful\/}
followed by a negative infinitival complement, as in the following
examples:
\begin{eg}{bookstrip-tc}
\item \ {\em To book the strip, fold the bottom third or more of the strip 
over the middle of the panel, pasted sides together, {\bf taking care
not to crease the wallpaper sharply at the fold.}}
\end{eg}
\begin{eg}{bc}
\item \m{If your plans call for replacing the wood base molding with vinyl
cove molding, {\bf be careful not to damage the walls} as you remove the
wood base.}
\end{eg}
\eite

The question of interest for us is under which conditions one or the
other of the surface forms is chosen.  We are currently using this
information to drive the generation of warning messages in the {\sc
drafter} system \cite{preventions-inlg96}.  We will start by
discussing previous work on negative imperatives, and by presenting an
hypothesis to be explored.  We will then describe the nature of our
corpus and our coding schema, detailing the results of our inter-coder
reliability tests.  Finally, we will describe the results of our
analysis of the correlation between function and form features.

\section{Related work on Negative Imperatives}

While instructional text has sparked much interest in both the
semantics/pragmatics community and the computational linguistics
community, little work on preventative expressions, and in particular
on negative imperatives, has been done.  This lack of interest in the
two communities has been in some sense complementary.

In semantics and pragmatics, negation has been extensively studied
(cf. Horn \shortcite{horn}).  Imperatives, on the other hand, have not
(for a notable exception, see Davies \shortcite{edavies}).

In computational linguistics, on the other hand, positive imperatives
have been extensively investigated, both from the point of view of
interpretation \cite{homer,floabn2,chapman,thesis} and generation
\cite{mellish89:generation,mckeown90:generation,drafter-ijcai95,vanlin95}.
Little work, however, has been directed at negative imperatives.  (for
exceptions see the work of \newcite{homer} in interpretation and of
\newcite{ansari95} in generation).

\section{A Priori Hypotheses}\label{a-priori-section}

Di Eugenio \shortcite{thesis} put forward the following hypothesis
concerning the realization of preventative expressions.  In this
discussion, S refers to the instructor (speaker / writer) who is
referred to with feminine pronouns, and H to the agent (hearer /
reader), referred to with masculine pronouns:
\begin{itemize}
\item {\bf DONT imperatives.} A \m{DONT}  imperative is used when 
S expects H to be \m{aware} of a certain choice point, but to be
likely to choose the \m{wrong} alternative among many --- possibly
infinite --- ones, as in:
\begin{eg}{clean-parquet}
\item \ 
\m{Dust-mop or vacuum your parquet floor as you would carpeting.
{\bf Do not scrub or wet-mop the parquet.}}
\end{eg}
\noindent
Here, H is aware of the choice of various cleaning methods, but may
choose an inappropriate one (i.e., scrubbing or wet-mopping).

\item
{\bf Neg-TC imperatives.} In general, \m{neg-TC} imperatives are used
when S expects H to \m{overlook} a certain choice point; such choice
point may be identified through a possible side effect that the wrong
choice will cause.  It may, for example, be used when H might execute
an action in an undesirable way.  Consider:
\begin{eg}{drill}
\item \m{To make a piercing cut, first drill a hole in the waste stock on the
interior of the pattern.  If you want to save the waste stock for later
use, drill the hole near a corner in the pattern. {\bf Be careful not
to drill through the pattern line.}}
\end{eg}

Here, H has some choices as regards the exact position where to drill,
so S constrains him by saying {\em Be careful not to drill through the
pattern line}.
\end{itemize}

So the hypothesis is that H's \m{awareness} of the presence of a
certain choice point in executing a set of instructions affects the
choice of one preventative expression over another.  This hypothesis,
however, was based on a small corpus and on intuitions.  In this paper
we present a more systematic analysis.

\section{Corpus and coding}

Our interest is in finding correlations between features related to
the \m{function} of a preventative expression, and those related to
the \m{form} of that expression.  Functional features are the semantic
features of the message being expressed and the pragmatic features of
the context of communication.  The form feature is the grammatical
structure of the expression.  In this section we will start with a
discussion of our corpus, and then detail the function and form
features that we have coded.  We will conclude with a discussion of
the inter-coder reliability of our coding.

\subsection{Corpus}\label{corpus}

The raw instructional corpus from which we take all the examples we
have coded has been collected opportunistically off the internet and
from other sources.  It is approximately 4 MB in size and is made
entirely of written English instructional texts.  The corpus includes
a collection of recipes (1.7 MB), two complete do-it-yourself manuals
\cite{digest,home-repair} (1.2 MB)\footnote{These do-it-yourself
manuals were scanned by Joseph Rosenzweig.}, a set of computer games
instructions, the Sun Open-windows on-line instructions, and a
collection of administrative application forms.  As a collection,
these texts are the result of a variety of authors working in a
variety of instructional contexts.

We broke the corpus texts into expressions using a simple sentence
breaking algorithm and then collected the negative imperatives by
probing for expressions that contain the grammatical forms we were
interested in (e.g., expressions containing phrases such as ``don't''
and ``take care'').  The first row in Table~\ref{neg-table} shows the
frequency of occurrence for each of the grammatical forms we probed
for.  These grammatical forms, 1175 occurrences in all, constitute
2.5\% of the expressions in the full corpus.  We then filtered the
results of this probe in two ways:

\begin{enumerate}
\item When the probe returned more than 100 examples for a grammatical
form, we randomly selected around 100 of those returned.  We took all
the examples for those forms that returned fewer than 100 examples.
The number of examples that resulted is shown in row 2 of
Table~\ref{neg-table} (labelled ``raw sample'').
\item We removed those examples that, although they contained the
desired lexical string, did not constitute negative imperatives.  This
pruning was done when the example was not an imperative (e.g., ``If
you {\bf don't} see the Mail Tool window \ldots'') and when the
example was not negative (e.g., ``Make sure to lock the bit tightly in
the collar.'').  The number of examples which resulted is shown in row
3 of Table~\ref{neg-table} (labelled ``final coding'').  Note that the
majority of the ``make sure'' examples were removed here because they
were ensurative.
\end{enumerate}

\noindent
As shown in Table~\ref{neg-table}, the final corpus sample is made up
of 239 examples, all of which have been coded for the features to be
discussed in the next two sections. 

\begin{table*}[t]
\centering
\begin{tabular}{||l||c|c||c|c|c|c||}\hline 
& \multicolumn{2}{c||}{DONT} & \multicolumn{4}{c||}{Neg-TC}  \\ \hline
& {don't} & {do not} & {take care} & {make sure} & {be careful} &
{be sure} \\ \hline \hline
Raw Grep & 417 & 385  & 21 & 229 & 52 & 71 \\ \hline 
Raw Sample & 100 & 99  & 21 & 104 & 52 & 71\\ \hline 
Final Coding & 78 & 89  & 17 & 3 & 46 & 6\\ \hline 
& \multicolumn{2}{c||}{167}  & \multicolumn{4}{c||}{72}  \\ \hline
\end{tabular}
\caption{Distribution of negative imperatives}
\label{neg-table}
\end{table*}

\subsection{Form}

Because of its syntactic nature, the form feature coding was very
robust.  The possible feature values were: {\bf DONT} --- for the
\m{do not} and \m{don't} forms discussed above;
and {\bf neg-TC} --- for \m{take
care},
\m{make sure}, \m{ensure}, \m{be careful}, \m{be sure}, \m{be certain}
expressions with negative arguments.

\subsection{Function Features}\label{function-features}

The design of semantic/pragmatic features usually requires a series of
iterations and modifications.  We will discuss our schema, explaining
the reasons behind our choices when necessary.  We coded for two
function features: {\sc intentionality} and {\sc awareness}, which we
will illustrate in turn using $\alpha$ to refer to the negated
action. The conception of these features was inspired by the
hypothesis put forward in Section~\ref{a-priori-section}, as we will
briefly discuss below.

\subsubsection{Intentionality}

This feature encodes whether the agent consciously adopts the
intention of performing $\alpha$.  We settled on two values,
CON(scious) and UNC(onscious).  As the names of these values may be
slightly misleading, we discuss them in detail here:

\bdes
\item[{\sc CON}] is used to code situations where S expects H to intend
to perform $\alpha$. This often happens when S expects H to be aware
that $\alpha$ is an alternative to the $\beta$ H should perform, and
to consider them equivalent, while S knows that this is not the case.
Consider Ex.~\refeg{clean-parquet} above. If the negative imperative
{\em Do not scrub or wet-mop the parquet} were not included, the agent
might have chosen to \m{scrub} or \m{wet-mop} because these actions
may result in deeper cleaning, and because he was unaware of the bad
consequences.

\item[{\sc UNC}] is perhaps a less felicitous name because we
certainly don't mean that the agent may perform actions while being
unconscious!  Rather, we mean that the agent doesn't realise that
there is a choice point
It is used in two situations: when $\alpha$ is totally
accidental, as in:
\begin{eg}{burn1}
\item {\em Be careful not to burn the garlic.}
\end{eg}

\noindent
In the domain of cooking, no agent would consciously burn the garlic.
Alternatively, an example is coded as UNC when $\alpha$ has to be
intentionally planned for, but the agent may not take into account a
crucial feature of $\alpha$, as in:

\begin{eg}{temperature}
\item {\em Don't charge -- or store --  a tool where 
the temperature is below 40 degrees F or above 105 degrees.}
\end{eg}
While clearly the agent will have to intend to perform \m{charging} or
\m{storing a tool},  he is likely to overlook, at least in S's conception,
that temperature could have a negative impact on the results of such
actions.
\edes

\subsubsection{Awareness}

This binary feature captures whether the agent is AWare or UNAWare
that the consequences of $\alpha$ are bad.   These features are
detailed now:

\bdes

\item[{\sc UNAW}] is used when H is perceived to be unaware that
$\alpha$ is bad. For example, Example~\refeg{temperature} (``Don't
charge -- or store -- a tool where the temperature is below 40 degrees
F or above 105 degrees'') is coded as UNAW because it is unlikely that
the reader will know about this restriction;

\item[{\sc AW}] is used when H is aware that $\alpha$ is bad.
Example~\refeg{burn1} (``Be careful not to burn the garlic'') is coded
as AW because the reader is well aware that burning things when
cooking them is bad.  

\edes

\subsection{Inter-coder reliability}
\label{inter}

Each author independently coded each of the features for all the
examples in the sample.  
The percentage agreement is 76.1\% for intentionality and 92.5\% for
awareness.  Until very recently, these values would most likely have
been accepted as a basis for further analysis.  To support a more
rigorous analysis, however, we have followed Carletta's suggestion
\shortcite{carletta96} of using the K coefficient \cite{siegel88} as a
measure of coder agreement.  This statistic not only measures
agreement, but also factors out chance agreement, and is used for
nominal (or categorical) scales.  In nominal scales, there is no
relation between the different categories, and classification induces
equivalence classes on the set of classified objects.  In our coding
schema, each feature determines a nominal scale on its own. Thus, we
report the values of the K statistics for each feature we coded for.

If $P(A)$ is the proportion of times the coders agree, and $P(E)$ is
the proportion of times that coders are expected to agree by chance, K
is computed as follows: $$ {K \: = \:
{\frac{P(A)\,-\,P(E)}{1\,-\,P(E)}}} $$ Thus, if there is total
agreement among the coders, K will be 1; if there is no agreement
other than chance agreement, K will be 0. There are various ways of
computing $P(E)$; according to
\newcite{siegel88}, most researchers agree on the following formula,
which we also adopted: $$P(E) \: = \: {\sum_{j=1}^{m} p_{j}^{2}} $$
where $m$ is the number of categories, and $p_{j}$ is the proportion
of objects assigned to category $j$.

The mere fact that K may have a value $k$ greater than zero is not
sufficient to draw any conclusion, though, as it must be established
whether $k$ is significantly different from zero.  While
\newcite[p.289]{siegel88} point out that it is possible to check the
significance of K when the number of objects is large,
\newcite{rietveld93} suggest a much simpler correlation between K values and
inter-coder reliability, shown in Figure~\ref{kappa-levels}.

\begin{table}[t]
\centering
\begin{tabular}{|| l | l ||} \hline
Kappa Value & Reliability Level \\ \hline
.00 --  .20 & slight \\
.21 --  .40 & fair \\
.41 --  .60 & moderate \\
.61 --  .80 & substantial \\
.81 --	 1.00 & almost perfect \\ \hline
\end{tabular}
\caption{The Kappa Statistic and Inter-coder Reliability}
\label{kappa-levels}
\end{table}

For the form feature, the Kappa value is 1.0, which is not surprising
given its syntactic nature.  The function features, which are more
subjective in nature, engender more disagreement among coders, as
shown by the K values in Table~\ref{function-kappa}.  According to
Rietveld and van Hout, the awareness feature shows ``substantial''
agreement and the intentionality feature shows ``moderate'' agreement.

\begin{table}[t]
\centering
\begin{tabular}{||l|c||}\hline
feature & K  \\ \hline 
{\sc intentionality} & 0.51 \\
{\sc awareness} & 0.75 \\ 
\hline
\end{tabular}
\caption{Kappa values for function features}
\label{function-kappa}
\end{table}

\section{Analysis}\label{analysis}

In our analysis, we have attempted to discover and to empirically
verify correlations between the function features and the form
feature.  We did this by computing $\chi^2$ statistics for the
various functional features as they compared with form distinction
between DONT and neg-TC imperatives.  Given that the features were all
two-valued we were able to use the following definition of the
statistic, taken from  \cite{siegel88}:

$$ \chi^2  = \frac{N(|AD-BC|-\frac{N}{2})^2}{(A+B)(C+D)(A+C)(B+D)}$$

\noindent
Here N is the total number of examples and A-D are the values of the
elements of the 2$\times$2 contingency table (see
Figure~\ref{intentionality-contingency-table}).  The $\chi^2$
statistic is appropriate for the correlation of two independent
samples of nominally coded data, and this particular definition of it
is in line with Siegel's recommendations for 2$\times$2 contingency
tables in which $N>40$ \cite[page 123]{siegel88}.  Concerning the
assumption of independence, while it is, in fact, possible that some of the
examples may have been written by a single author,  the corpus was
written by a considerable number of authors.  Even the larger works
(e.g., the cookbooks and the do-it-yourself manuals) are collections of
the work of multiple authors.  We felt it acceptable, therefore, to
view the examples as independent and use the $\chi^2$ statistic.

To compute $\chi^2$ for the coded examples in our corpus, we collected
all the examples for which we agreed on both of the functional
features (i.e., intentionality and awareness).  Of the 239 total
examples, 165 met this criteria.  Table~\ref{significance} lists the
$\chi^2$ statistic and its related level of significance for each of
the features.  The significance levels for intentionality and
awareness indicate that the features do correlate with the forms.  We
will focus on these features in the remainder of this section.

\begin{table}[t]
\centering
\begin{tabular}{||l|c|c||}\hline
feature & $\chi^2$ & significance level \\ \hline 
intentionality & 51.4 & 0.001 \\
awareness & 56.9 & 0.001 \\
\hline
\end{tabular}
\caption{$\chi^2$ statistic and significance levels}
\label{significance}
\end{table}

The 2$\times$2 contingency table from which the intentionality value
was derived is shown in Table~\ref{intentionality-contingency-table}.
This table shows the frequencies of examples marked as conscious or
unconscious in relation to those marked as DONT and neg-TC.  A strong
tendency is indicated to prevent actions the reader is likely to
consciously execute using the DONT form.  Note that the table entry
for conscious/neg-TC is 0, indicating that there were no examples
marked as both CON and neg-TC.  Similarly, the neg-TC form is more
likely to be used to prevent actions the reader is likely to execute
unconsciously.

\begin{table}[t]
\centering
\begin{tabular}{l | c c | c}
& Conscious & Unconscious & \m{Total} \\ \hline
DONT & 61 (A) & 45 (B) & 106 \\ 
neg-TC & 0 (C) & 59 (D) & 59 \\ \hline
\m{Total} & 61 & 104 & 165 (N) \\ 
\end{tabular}
\caption{Contingency Table for Intentionality}
\label{intentionality-contingency-table}
\end{table}

In Section~\ref{a-priori-section} we speculated that the hearer's
awareness of the choice point, or more accurately, the writer's view
of the hearer's awareness, would affect the appropriate form of
expression of the preventative expression. In our coding, awareness
was then shifted to awareness of bad consequences rather than of
choices per se. However, the basic intuition that awareness plays a
role in the choice of surface form is supported, as the contingency
table for this feature in Table~\ref{awareness-contingency-table}
shows.  It indicates a strong preference for the use of the DONT form
when the reader is presumed to be unaware of the negative consequences
of the action to be prevented, the reverse being true for the use of
the neg-TC form.

\begin{table}[t]
\centering
\begin{tabular}{l | c c | c} 
& Aware & Unaware & \m{Total} \\ \hline
DONT & 3 & 103 & 106 \\ 
neg-TC & 32 & 27 & 59 \\ \hline
\m{Total} & 35 & 130 & 165 \\ 
\end{tabular}
\caption{Contingency Table for Awareness}
\label{awareness-contingency-table}
\end{table}

The results of this analysis, therefore, demonstrate that the
intentionality and awareness features do co-vary with grammatical
form, and in particular, support a form of the hypothesis put forward
in Section~\ref{a-priori-section}.

\section{Application}

We have successfully used the correlations discussed here to support
the generation of warning messages in the {\sc drafter} project
\cite{drafter-ieee96}.  {\sc drafter} is a technical authoring
support tool which generates instructions for graphical interfaces.
It allows its users to specify a procedure to be expressed in
instructional form, and in particular, allows them to specify actions
which must be prevented at the appropriate points in the procedure.
At generation time, then, {\sc drafter} must be able to select the
appropriate grammatical form for the preventative expression.

We have used the correlations discussed in this paper to build the
text planning rules required to generate negative imperatives.  This
is discussed in more detail elsewhere \cite{preventions-inlg96}, but
in short, we input our coded examples to Quinlan's C4.5 learning
algorithm \cite{quinlan93:other}, which induces a decision tree
mapping from the functional features to the appropriate form.
Currently, these features are set manually by the user as they are too
difficult to derive automatically.

\section{Conclusions}\label{df-summary}

This paper has detailed a corpus study of preventative expressions in
instructional text.  The study highlighted correlations between
functional features and grammatical form, the sort of correlations
useful in both interpretation and generation.  Studies such as this
have been done before in Computational Linguistics, although not, to
our knowledge, on preventative expressions.  The point we want to
emphasise here is a methodological one.  Only recently have studies
been making use of more rigorous statistical measures of accuracy and
reproducibility used here.  We have found the Kappa statistic critical
in the definition of the features we coded (see Section~\ref{inter}).

We intend to augment and refine the list of features discussed here
and hope to use them in understanding applications as well as
generation applications.  We also intend to extend the analysis to
ensurative expressions.

\bibliographystyle{acl}

\end{document}